\documentclass{article}

\usepackage[preprint]{nips_2018}

\usepackage{graphicx}

\usepackage[utf8]{inputenc} % allow utf-8 input
\usepackage[T1]{fontenc}    % use 8-bit T1 fonts
\usepackage{hyperref}       % hyperlinks
\usepackage{url}            % simple URL typesetting
\usepackage{booktabs}       % professional-quality tables
\usepackage{amsfonts}       % blackboard math symbols
\usepackage{nicefrac}       % compact symbols for 1/2, etc.
\usepackage{microtype}      % microtypography
\usepackage{amsmath,amssymb}
\usepackage{mathtools}
\usepackage{enumerate}
\usepackage{enumitem}
\usepackage{bm}

\DeclareMathOperator*{\argmax}{arg\,max}

\title{A0C: Alpha Zero in Continuous Action Space}

\author{
  Thomas M. Moerland$^{\ast\dagger}$, Joost Broekens$^{\ast}$, Aske Plaat$^{\dagger}$ and Catholijn M. Jonker$^{\ast\dagger}$ \\
   \\
 $^\ast$Dep. of Computer Science, Delft University of Technology, The Netherlands \\ 
 $^\dagger$Dep. of Computer Science, Leiden University, The Netherlands
}

\begin{document}

\maketitle

\begin{abstract}
A core novelty of Alpha Zero is the interleaving of tree search and deep learning, which has proven very successful in board games like Chess, Shogi and Go. These games have a discrete action space. However, many real-world reinforcement learning domains have continuous action spaces, for example in robotic control, navigation and self-driving cars. This paper presents the necessary theoretical extensions of Alpha Zero to deal with continuous action space. We also provide some preliminary experiments on the Pendulum swing-up task, empirically showing the feasibility of our approach. Thereby, this work provides a first step towards the application of iterated search and learning in domains with a continuous action space.
\end{abstract}.

\section{Introduction} \label{sec_introduction}
\begin{figure}[t]
  \centering
      \includegraphics[width=0.5\textwidth]{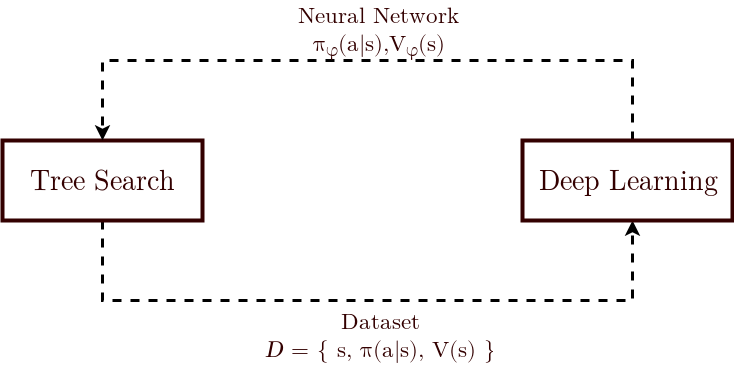}
  \caption{Iterated tree search and function approximation.}
    \label{fig_searchrl}
\end{figure}

Alpha Zero has achieved state-of-the-art, super-human performance in Chess, Shogi \citep{silver2017mastering2} and the game of Go \citep{silver2016mastering,silver2017mastering}. The key innovation of Alpha Zero compared to traditional reinforcement learning approaches is the use of a small, nested tree search as a policy evaluation.\footnote{Additionally, the tree search provides an efficient exploration method, which is a key challenge in reinforcement learning \cite{moerland2017efficient}.} Whereas traditional reinforcement learning treats each environment step or trace as an individual training target, Alpha Zero aggregates the information of multiple traces in a tree, and eventually aggregates these tree statistics into targets to train a neural network. The neural network is then used as a prior to improve new tree searches. This closes the loop between search and function approximation (Figure \ref{fig_searchrl}). In section \ref{sec_dicussion} we further discuss why this works so well. 

While Alpha Zero has been very successful in two-player games with discrete action spaces, it is not yet applicable in continuous action space (nor has Alpha Zero been tested in single-player environments). Many real-world problems, such as robotics control, navigation and self-driving cars, have a continuous action space. We will now list the core contributions of this paper. Compared to the Alpha Zero paradigm for discrete action spaces, we require:
\begin{enumerate}
\item A Monte Carlo Tree Search (MCTS) method that works in continuous action space. We built here on earlier results on {\it progressive widening} (Section \ref{sec_progressive_widening}). 
\item Incorporation of a continuous prior to steer a new MCTS iteration. While Alpha Zero uses the discrete density as a prior in a (P)UCT formula \citep{rosin2011multi,kocsis2006bandit}, we need to leverage a continuous density (which is unbounded) to direct the next MCTS iteration (Section \ref{sec_mcts_prior}) 
\item A training method. Alpha Zero transforms the MCTS visitation counts to a discrete probability distribution. We need to estimate a continuous density from a set of support points, and specify an appropriate training loss in continuous policy space (Section \ref{sec_loss}).
\end{enumerate}

The remainder of this paper is organized as follows. Section \ref{sec_preliminaries} presents essential preliminaries on reinforcement learning and MCTS. Section \ref{sec_mcts} discusses the required MCTS modifications for a continuous action space with a continuous prior (Fig. \ref{fig_searchrl}, upper part of the loop). In Section \ref{sec_loss} we cover the generation of training targets from the tree search and specify an appropriate neural network loss (Fig. \ref{fig_searchrl}, lower part of the loop). Sections \ref{sec_experiments}, \ref{sec_dicussion} and \ref{sec_conclusion} present experiments, discussion and conclusions. 

\section{Preliminaries} \label{sec_preliminaries}
\paragraph{Markov Decision Process} We adopt a finite-horizon Markov Decision Process (MDP) \citep{sutton2018reinforcement} given by the tuple $\{\mathcal{S},\mathcal{A},f,\mathcal{R},\gamma,T\}$, where $\mathcal{S} \subseteq \mathbb{R}^{n_s}$ is a state set, $\mathcal{A} = \subseteq \mathbb{R}^{n_a}$ continuous action set, $f: \mathcal{S} \times \mathcal{A} \to P(\mathcal{S})$ denotes a transition function, $\mathcal{R}: \mathcal{S} \times \mathcal{A} \to \mathbb{R}$ a (bounded) reward function, $\gamma \in (0,1]$ a discount parameter and $T$ the time horizon. At every time-step $t$ we observe a state $\bm{s}^t \in \mathcal{S}$ and pick an action $\bm{a}^t \in \mathcal{A}$, after which the environment returns a reward $r^t = \mathcal{R}(\bm{s}^t,\bm{a}^t)$ and next state $\bm{s}^{t+1} = f(\bm{s}^t,\bm{a}^t)$. We act in the MDP according to a stochastic policy $\pi: \mathcal{S} \to P(\mathcal{A})$. Define the (policy-dependent)  state value $V^\pi(\bm{s}^t) = \mathrm{E}_\pi[ \sum_{k=0}^T (\gamma)^k \cdot r^{t+k}]$ and state-action value $Q^\pi(\bm{s}^t,\bm{a}^t) = \mathrm{E}_\pi[ \sum_{k=0}^T (\gamma)^k \cdot r^{t+k}]$, respectively. Our goal is to find a policy $\pi$ that maximizes this cumulative, discounted sum of rewards.

\paragraph{Monte Carlo Tree Search} We present a brief introduction of the well-known MCTS algorithm \citep{coulom2006efficient,browne2012survey}. In particular, we discuss a variant of the PUCT algorithm \citep{rosin2011multi}, as also used in Alpha Zero \citep{silver2017mastering2,silver2017mastering}. Every action node in the tree stores statistics $\{n(\bm{s},\bm{a}),W(\bm{s},\bm{a}),Q(\bm{s},\bm{a})\}$, where $n(\bm{s},\bm{a})$ is the visitation count, $W(\bm{s},\bm{a})$ the cumulative return over all roll-outs through $(\bm{s},\bm{a})$, and $Q(\bm{s},\bm{a}) = W(\bm{s},\bm{a})/n(\bm{s},\bm{a})$ is the mean action value estimate. PUCT alternates four phases:
\begin{enumerate}
\item {\bf Select}
In the first stage, we descent the tree from the root node according to:

\begin{equation}
\pi_{tree}(\bm{a}|\bm{s}) = \argmax_{\bm{a}} \Bigg[ Q(\bm{s},\bm{a}) + c_{\text{puct}} \cdot \pi_\phi(\bm{a}|\bm{s}) \cdot \frac{\sqrt{n(\bm{s})}}{n(\bm{s},\bm{a}) + 1} \Bigg] \label{eq_puct_alphazero}
\end{equation}

where $n(\bm{s}) = \sum_{\bm{a}} n(\bm{s},\bm{a})$ is the total number of visits to state $s$ in the tree, $c_{\text{puct}} \in \mathbb{R}^+$ is a constant that scales the amount the exploration/optimism, and $\pi_\phi(\bm{a}|\bm{s})$ is the probability assigned to action $\bm{a}$ by the network.\footnote{This equation differs from the standard UCT-like formulas in two ways. The $\pi_\phi(a|s)$ term scales the confidence interval based on prior knowledge, as stored in the the policy network. The  $+\hspace{0.05cm}1$ term in the denominator ensures that the policy prior already affects the decision when there are unvisited actions. Otherwise, every untried action would be tried at least once, since without the $+1$ term Eq. \ref{eq_alphazero_policy} becomes $\infty$ for untried actions. This is undesirable for large action spaces and small trees, where we directly want to prune the actions that we already know are inferior from prior experience.} The tree policy is followed until we either reach a terminal state or select an action we have not tried before.

\item {\bf Expand} We next expand the tree with a new leaf state $\bm{s}_L$\footnote{We use superscript $\bm{s}^t$ to index real environment states and actions, subscripts $\bm{s}_d$ to index states and actions at depth $d$ in the search tree, and double subscripts $\bm{a}_{d,j}$ to index a specific child action $j$ at depth $d$. For example, $\bm{a}_{0,0}$ is the first child action at the root $\bm{s}_0$. At every timestep $t$, the tree root $\bm{s}_0:=\bm{s}^t$, i.e. the current environment state becomes the tree root.} obtained from simulating the environment with the new action from the last state in the current tree.

\item {\bf Roll-out} We then require an estimate of the value $V(\bm{s}_L)$ of the new leaf node, for which MCTS uses the sum of reward of a (random) roll-out $\mathrm{R}(\bm{s}_L)$. In Alpha Zero, this gets replaced by the prediction of a value network $\mathrm{R}(\bm{s}_L) := V_\phi(\bm{s}_L)$. 

\item {\bf Back-up} Finally, we recursively back-up the results in the tree nodes. Denote the current forward trace in the tree as $\{\bm{s}_0,\bm{a}_0,\bm{s}_1, .. \bm{s}_{L-1},\bm{a}_{L-1},\bm{s}_L\}$. Then, for each state-action edge $(\bm{s}_i,\bm{a}_i)$, $L > i \geq 0$, we recursively estimate the state-action value as

\begin{equation}
\mathrm{R}(\bm{s}_i,\bm{a}_i) = r(\bm{s}_i,\bm{a}_i) + \gamma \mathrm{R}(\bm{s}_{i+1},\bm{a}_{i+1}). \label{eq_rollout}
\end{equation} 

where $\mathrm{R}(\bm{s}_L,\bm{a}_L) := \mathrm{R}(\bm{s}_L)$. We then increment $W(\bm{s}_i,\bm{a}_i)$ with the new estimate $\mathrm{R}(\bm{s}_i,\bm{a}_i)$, increment the visitation count $n(\bm{s}_i,\bm{a}_i)$ with 1, and set the mean estimate to $Q(\bm{s}_i,\bm{a}_i) = W(\bm{s}_i,\bm{a}_i)/n(\bm{s}_i,\bm{a}_i)$. We repeatedly apply this back-up one step higher in the tree until we reach the root node $s_0$. 

\end{enumerate}

This procedure is repeated until the overall MCTS trace budget $N_{\text{trace}}$ is reached. MCTS returns a set of root actions $ A_0 = \{ \bm{a}_{0,0},\bm{a}_{0,1},..,\bm{a}_{0,m} \}$ with associated counts $ N_0 =  \{n(\bm{s}_0,\bm{a}_{0,0}),n(\bm{s}_0,\bm{a}_{0,1}), .., n(\bm{s}_0,\bm{a}_{0,m})\}$. Here $m$ denotes the number of child actions, which for Alpha Zero is always fixed to the cardinality of the discrete action space $m=|\mathcal{A}|$. We select the real action $\bm{a}^t$ to play in the environment by sampling from the probability distribution obtained from normalizing the action counts at the root $\bm{s}_0 (=\bm{s}^t)$:

\begin{equation}
\bm{a}^t \sim \hat{\pi}(\bm{a}|\bm{s}_0), \quad \text{where} \quad \hat{\pi}(\bm{a}|\bm{s}_0) = \frac{n(\bm{s}_0,\bm{a})}{n(\bm{s}_0)} \label{eq_alphazero_policy}
\end{equation}

and $n(\bm{s}_0) = \sum_{\bm{b} \in A_0} n(\bm{s}_0,\bm{b})$. Note that $n(\bm{s}_0) \geq N_{\text{trace}}$, since we store the subtree that belongs to the picked action $\bm{a}^t$ for the MCTS at the next timestep. 

\paragraph{Neural Networks} We introduce two neural networks - similar to Alpha Zero - to estimate a parametrized policy $\pi_\phi(\bm{a}|\bm{s})$ and the state value $V_\phi(\bm{s})$. Both networks share the initial layers. The joint set of parameters of both networks is denoted by $\phi$. The neural networks are trained on targets generated by the MCTS procedure. These training targets, extracted from the tree search, are denoted by $\hat{\pi}(\bm{a}|\bm{s})$ and $\hat{V}(\bm{s})$.

\section{Tree Search in Continuous Action Space} \label{sec_mcts}
As noted in the introduction, we require two modifications to the MCTS procedure: 1) a method to deal with continuous action spaces, and 2) a way to include a continuous policy network into the MCTS search. 

\subsection{Progressive Widening} \label{sec_progressive_widening}
During MCTS with a discrete action space we evaluate the PUCT formula for {\it all} actions. However, in continuous action space we can not enumerate all actions, i.e., there are actually infinitely many actions in a continuous set. A practical solution to this problem is {\it progressive widening} \citep{coulom2007computing,chaslot2008progressive}, where we make the number of child actions of state $\bm{s}$ in the tree $m(\bm{s})$ a function of the total number of visits to that state $n(\bm{s})$. This implies that actions with good returns, which will get more visits, will also gradually get more child actions for consideration. In particular, \citet{couetoux2011continuous} uses 

\begin{equation}
m(\bm{s}) = c_{pw} \cdot n(\bm{s})^\kappa
\end{equation}

for constants $c_{pw} \in \mathbb{R}^+$ and $\kappa \in (0,1)$, making $m(\bm{s})$ a polynomial (root) function of $n(\bm{s})$. The idea of progressive widening was introduced by \citet{coulom2007computing}, who made $m(\bm{s})$ a logarithmic function of $n(\bm{s})$. Although originally conceived for discrete domains, this technique turns out to be an effective solution for continuous action space as well \citep{couetoux2011continuous}. 

\subsection{Continuous policy network prior} \label{sec_mcts_prior}
For now assume we manage to train a policy network $\pi_\phi(\bm{s})$ from the results of the MCTS procedure. Alpha Zero can enumerate the probability for all available discrete actions, and uses this probability as a prior scaling on the upper confidence bound term in the UCT formula (Eq. \ref{eq_puct_alphazero}). For the continuous policy space, we could use a similar equation, where we use $\pi_\phi(\bm{a}|\bm{s})$ of the considered $\bm{a}$ as predicted by the network. However, the continuous $\pi_\phi(\bm{a}|\bm{s})$ is unbounded.\footnote{For a discrete probability distribution, $\pi(\bm{a}) \leq 1 \hspace{0.1cm} \forall \bm{a}$. However, although the probability density function (pdf) of continuous random variables integrates to 1, i.e. $\int \pi(\bm{a}|\bm{s}) \mathrm{d} \bm{a} = 1$, this does not bound the value of the pdf $\pi(\bm{a})$ at a particular point $\bm{a}$, i.e. $\pi(\bm{a}) \in [0,\infty)$.} This gives us the risk of rescaling/stretching the confidence intervals too much. Another option - which we consider in this work - is to use the policy network to sample new child actions in the tree search (when adding a new action based on progressive widening). Thereby, the policy net steers the actions that we will consider in the tree search. This has a similar effect as Eq. \ref{eq_puct_alphazero} for AlphaGo Zero does, as it effectively prunes away child actions in subtrees of which we already know that they perform poorly. 

\section{Neural network training in continuous action space} \label{sec_loss}
We next want to use the MCTS output to improve our neural networks. Compared to Alpha Zero, the continuous action space forces us to come up with a different policy network specification, policy target calculation and training loss. These aspects are covered in Section \ref{sec_policy}. Afterwards, we briefly detail the value network training procedure, including a slight variant of the value target estimation (Section \ref{sec_value}). 

\subsection{Policy Network} \label{sec_policy}
\paragraph{Policy Network Distribution}
We require a neural network that outputs a continuous density. However, continuous action spaces usually have some input bounds. For example, when we learn the torques or voltages on a robot manipulator, then a too extreme torque/voltage may break the motor altogether. Therefore, continuous actions spaces are generally symmetrically bounded to some $[-c_b,c_b]$ interval, for scalar $c_b \in \mathbb{R}^+$. To ensure that our density predicts in this range, we use a transformation of a factorized Beta distribution $\pi_\phi(\bm{a}|\bm{s}) = g(\bm{u})$, with elements $u_i \sim \text{Beta}(\alpha_i(\phi),\beta_i(\phi))$ and deterministic transformation $g(\cdot)$. Details are provided in Appendix \ref{sec_bounds}. Note that the remainder of this section holds for any $\pi_\phi(\bm{a}|\bm{s})$ network output distribution from which we know how to sample and evaluate (log) densities. 

\paragraph{Training Target} We want to transform the result of the MCTS with progressive widening to a continuous target density $\hat{\pi}$ (to training our neural network with). Recall that MCTS returns the sets $A_0$ and $N_0$ of root actions and root counts, respectively. We can not normalize these counts like Alpha Zero does (Eq. \ref{eq_alphazero_policy}) for the discrete case. The only assumption, similar to Alpha Zero, that we make here is that the density at a root action $\bm{a}_{0,i}$ is proportional to the visitation counts, i.e.\footnote{The remainder of this section always concerns the root state $\bm{s}_0$ and root actions $\bm{a}_{0,i}$. Therefore, we omit the depth subscript (of $0$) for readability.}

\begin{equation}
\hat{\pi}(\bm{a}_{i}|\bm{s})  = \frac{n(\bm{s},\bm{a}_i)^\tau}{Z(\bm{s},\tau)} \label{eq_pi_unnormalized}
\end{equation}

where $\tau \in \mathbb{R}^+$ specifies some temperature parameter, and $Z(\bm{s},\tau)$ is a normalization term (that is assumed to not depend on $\bm{a}_i$, as the density at the support points is only proportional to the counts). Note that this does not define a proper density, as we never specified a density in between the support points. However, we can ignore this issue, as we will only consider the loss at the support points.

\paragraph{Loss}
In short, our main idea is to leave the normalization and generalization of the policy over the action space to the network loss. If we specify a network output distribution that enforces $\int_a \pi_\phi(\bm{a}|\bm{s}) = 1$, i.e., making it a proper continuous density, then we may specify a loss with respect to a target density $\hat{\pi}(\bm{a}|\bm{s})$, {\it even when the target density is only known on a relative scale}. More extreme counts (relative densities) will produce stronger gradients, and the restrictions of the network output density will ensure that we can not pull the density up or down over the entire support (as it needs to integrate to 1). This way, we make our network output density mimic the counts on a relative scale. 

We will first give a general derivation, acting as if $\hat{\pi}(\bm{a}|\bm{s})$ is a proper density, and swap in the empirical density at the end. We minimize a policy loss $\mathcal{L}^\text{policy}(\phi)$ based on the Kullback-Leibler divergence between the network output $\pi_\phi(\bm{a}|\bm{s})$ and the empirical density $\hat{\pi}(\bm{a}|\bm{s})$ (Eq. \ref{eq_pi_unnormalized}):

\begin{equation}
\mathcal{L}^\text{policy}(\phi) = \text{D}_{KL} \Big( \pi_\phi(\bm{a}|\bm{s}) \| \hat{\pi}(\bm{a}|\bm{s}) \Big)  =  \mathrm{E}_{a \sim \pi_\phi(\bm{a}|\bm{s})} \Big[ \log \pi_\phi(\bm{a}|\bm{s}) - \log \hat{\pi}(\bm{a}|\bm{s}) \Big] \label{eq_policy_loss}
\end{equation}

We may use the REINFORCE\footnote{The REINFORCE trick \citep{williams1992simple}, also known as the likelihood ratio estimator, is an identity regarding the derivative of an expectation, when the expectation depends on the parameter towards which we differentiate: $\nabla_\phi \mathrm{E}_{\bm{a} \sim p_\phi(\bm{a})} [ f(\bm{a}) ] = \mathrm{E}_{\bm{a} \sim p_\phi(\bm{a})} [ f(\bm{a}) \nabla_\phi \log p_\phi(\bm{a})]$, for some function $f(\cdot)$ of $\bm{a}$.} trick to get an unbiased gradient estimate of the above loss:

\begin{align}
\nabla_\phi \mathcal{L}^\text{policy}(\phi) &= \nabla_\phi \mathrm{E}_{a \sim \pi_\phi(\bm{a}|\bm{s})} \Big[ \log \pi_\phi(\bm{a}|\bm{s}) - \tau \log n(\bm{a},\bm{s}) + \log Z(\bm{s},\tau)\Big] \nonumber \\ 
&= \mathrm{E}_{a \sim \pi_\phi(\bm{a}|\bm{s})} \Big[  \Big(  \log \pi_\phi(\bm{a}|\bm{s}) - \tau \log n(\bm{a},\bm{s}) + \log Z(\bm{s},\tau) \Big) \nabla_\phi \log \pi_\phi(\bm{a}|\bm{s}) \Big] \nonumber 
\end{align}

We now drop $Z(\bm{s},\tau)$ since it does not depend on $\phi$ (or chose an appropriate state-dependent baseline, as is common with REINFORCE estimators). Moreover, we replace the expectation over $\bm{a} \sim \pi_\phi(\bm{a}|\bm{s})$ with the empirical support points $\bm{a}_i \sim \mathcal{D}_{\bm{s}}$, where $\mathcal{D}_{\bm{s}}$ denotes the subset of the database containing state $\bm{s}$. Our final gradient estimator becomes

\begin{align}
\nabla_\phi \mathcal{L}^\text{policy}(\phi) = \mathrm{E}_{\bm{s} \sim \mathcal{D},\bm{a}_i \sim \mathcal{D}_s} \Big[  \Big(  \log \pi_\phi(\bm{a}_i|\bm{s}) - \tau \log n(\bm{s},\bm{a}_i) \Big) \nabla_\phi \log \pi_\phi(\bm{a}_i|\bm{s}) \Big] \label{eq_policy_gradients}
\end{align}

\paragraph{Entropy regularization}
Continuous policies have a risk to collapse \citep{haarnoja2018soft}. If all sampled actions are close to each other, then the distribution may narrow too much, loosing any exploration. In the worst case, the distribution may completely collapse, which will produce NaNs and break the training process. As we empirically observed this problem, we augment the training objective with an entropy maximization term. This prevents the policy from collapsing, and additionally ensures a minimum level of exploration. We define the entropy loss as

\begin{equation}
\mathcal{L}^H(\phi) = H(\pi_\phi(\bm{a}|\bm{s})) = - \int \pi_\phi(\bm{a}|\bm{s}) \log \pi_\phi(\bm{a}|\bm{s}) \mathrm{d}a.
\end{equation}

Details on the computation of the entropy for the case where $\pi_\phi(\bm{a}|\bm{s})$ is a transformed Beta distribution are provided in Appendix \ref{sec_entropy_beta}. The full policy loss thereby becomes

\begin{equation}
\mathcal{L}^{\pi}(\phi) = \mathcal{L}^\text{policy}(\phi) - \lambda \mathcal{L}^H(\phi),
\end{equation}

where $\lambda$ is a hyperparameter that scales the contribution of the entropy term to the overall loss.

\subsection{Value Network} \label{sec_value}
Value network training is almost identical to the Alpha Zero specification. The only thing we modify is the estimation of $\hat{V}(\bm{s})$, the training target for the value. Alpha Zero uses the eventual return of the full episode as the training target for every state in the trace. This is an unbiased, but high-variance signal (in reinforcement learning terminology \citep{sutton2018reinforcement}, it uses a full Monte Carlo target). Instead, we use the MCTS procedure as a value estimator, leveraging the action value estimates $Q(s_0,a)$ at the root $s_0$. We could weigh these according to the visitation counts at the root. However, we usually built relatively small trees,\footnote{AlphaGo Zero uses 1600 traces per timestep. We evaluate on smaller domains, and have less computational resources.} for which a non-negligible fraction of the traces are exploratory. Therefore, we propose an {\it off-policy} estimate of the value at the root:

\begin{equation}
\hat{V}(\bm{s}_0) = \max_{\bm{a}} Q(\bm{s}_0,\bm{a})
\end{equation}

The value loss $\mathcal{L}^V(\phi)$ is a standard mean-squared error loss: 

\begin{equation}
\mathcal{L}^V(\phi) = \mathrm{E}_{\bm{s} \sim \mathcal{D}} \Bigg[\Big(V_\phi(\bm{s}) - \hat{V}(\bm{s})\Big)^2 \Bigg]. 
\end{equation}

\begin{figure}[b]
  \centering
      \includegraphics[width=0.5\textwidth]{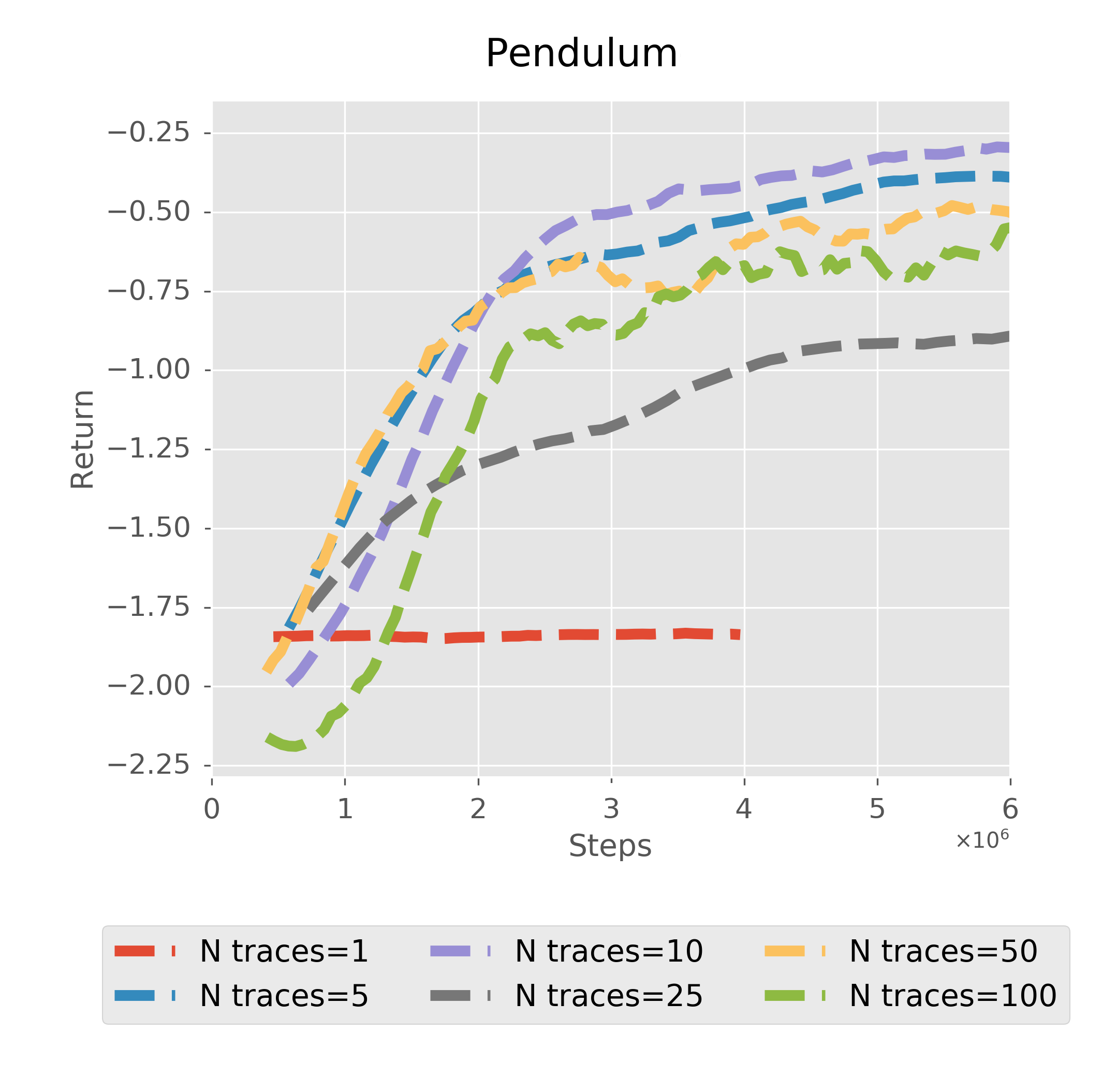}
  \caption{Learning curves for Pendulum domain. Compared to the OpenAI Gym implementation we rescale every reward by a factor $1/1000$ (which leaves the task and optimal solution unchanged). Results averaged over 10 repetitions.}
    \label{fig_pendulum}
\end{figure}

\section{Experiments} \label{sec_experiments}
Figure \ref{fig_pendulum} shows the results of our algorithm on the Pendulum-v0 task from the OpenAI Gym \citep{brockman2016openai}. The curves show learning performance for different computational budgets per MCTS at each timestep. Note that the x-axis displays true environment steps, which includes the MCTS simulations. For example, if we use 10 traces per MCTS, then every real environment step counts as 10 on this scale. 

First, we observe that our continuous Alpha Zero version does indeed learn on the Pendulum task. Interestingly, we observe different learning performance for different tree sizes, where the `sweet spot' appears to be at an intermediate tree size (of 10). For larger trees, we complete less episodes (a single episode takes longer) and therefore train our neural network less frequently. Therefore, although each individual trace gets more budget, it takes longer before the tree search starts to profit from improved network estimates (generalization). 

We train our neural network after every completed episode. However, the runs with smaller tree sizes complete much more episodes compared to the runs with a larger tree size. Moreover, the data generated from larger tree searches could be deemed `more trustworthy', as we spend more computational effort in generating them. We try to compensate for this effect by making the number of training epochs over the database after each episode proportional to the size of the nested tree search. Specifically, after each episode we train for

\begin{equation}
n_\text{epochs} = \Bigg \lceil \frac{N_\text{traces}}{c_e} \Bigg \rceil \label{eq_ceil}
\end{equation}

for constant $c_e \in \mathbb{R}^+$ and $\lceil \cdot \rceil$ denoting the ceiling function. In our experiments we set $c_e=20$. This may explain why the run with $N_\text{traces}=25$ performs suboptimal compared to the others, as the non-linearity in Eq. \ref{eq_ceil} (due to the ceiling function) may accidentally turn out bad for this number of tree traces. Moreover, note that the learning curve of training with a tree size of 1 is shorter than the other curves. This happens because we gave each run an equal amount of wall-clock time. The run with tree size 1 finishes much more episodes, and because $c_e > 1$ it still trains more frequently than the other runs, which makes it eventually perform less total steps in the domain.

\paragraph{Implementation details}
We use a three layer neural network with 128 units in each hidden layer and ELu activation functions. For the MCTS we set $c_{\text{puct}}=0.05$, $c_{\text{pw}}=1$ and $\kappa=0.5$, and for the policy loss $\lambda=0.1$ and $\tau=0.1$. We train the networks in Tensorflow \citep{abadi2016tensorflow}, using RMSProp optimizer on mini-batches of size 32 with a learning rate of 0.0001. Episodes last at maximum 300 steps.   

\section{Discussion} \label{sec_dicussion}
The results in Fig. \ref{fig_pendulum} reveal an interesting trade-off in the iterated tree search and function approximation paradigm. We hypothesize that the strength of tree search is the in the locality of information. Each edge stores its own statistics, and this makes it easy to locally separate the effect of actions. Moreover, the forward search gives a more stable value estimate, smoothing out local errors in the value network. In contrast, the strength of the neural network is generalization. Frequently, we re-encounter the (almost) same state in a different subtree during a next episode. Supervised learning is a natural way to generalize the already learned knowledge from a previous episode.

One of the key observations of the present paper is that we actually need both. If we {\it only} perform tree search, then we eventually fail at solving the domain because all information is kept locally. In contrast, if we only build trees of size 1, then we are continuously generalizing without ever locally separating decisions and improving our training targets. Our results suggest that there is actually a sweet spot halfway, where we build trees of moderate size, after which we perform a few epochs of training.

Future work will test the A0C algorithm in more complicated, continuous action space tasks \citep{brockman2016openai,todorov2012mujoco}. Moreover, our algorithm could profit from recent improvements in the MCTS algorithm \citep{moerland2018monte} and other network architectures \citep{szegedy2015going}, as also leveraged in Alpha Zero. 

\section{Conclusion} \label{sec_conclusion}
This paper introduced Alpha Zero for Continuous action space (A0C). Our method learns a continuous policy network - based on transformed Beta distributions - by minimizing a KL-divergence between the network distribution and an unnormalized density at the support points from the MCTS search. Moreover, the policy network also directs new MCTS searches by proposing new candidate child actions in the search tree. Preliminary results on the Pendulum task show that our approach does indeed learn. Future work will further explore the empirical performance of A0C. In short, A0C may be a first step in transferring the success of iterated search and learning, as observed in two-player board games with discrete action spaces \citep{silver2017mastering2,silver2017mastering}, to the single-player, continuous action space domains, like encountered in robotics, navigation and self-driving cars. 

\bibliographystyle{apalike}
\bibliography{example}

\begin{thebibliography}{}

\bibitem[Abadi et~al., 2016]{abadi2016tensorflow}
Abadi, M., Barham, P., Chen, J., Chen, Z., Davis, A., Dean, J., Devin, M.,
  Ghemawat, S., Irving, G., Isard, M., et~al. (2016).
\newblock {TensorFlow: A System for Large-Scale Machine Learning.}
\newblock In {\em {OSDI}}, volume~16, pages 265--283.

\bibitem[Brockman et~al., 2016]{brockman2016openai}
Brockman, G., Cheung, V., Pettersson, L., Schneider, J., Schulman, J., Tang,
  J., and Zaremba, W. (2016).
\newblock {Openai gym}.
\newblock {\em arXiv preprint arXiv:1606.01540}.

\bibitem[Browne et~al., 2012]{browne2012survey}
Browne, C.~B., Powley, E., Whitehouse, D., Lucas, S.~M., Cowling, P.~I.,
  Rohlfshagen, P., Tavener, S., Perez, D., Samothrakis, S., and Colton, S.
  (2012).
\newblock {A survey of monte carlo tree search methods}.
\newblock {\em IEEE Transactions on Computational Intelligence and AI in
  games}, 4(1):1--43.

\bibitem[Chaslot et~al., 2008]{chaslot2008progressive}
Chaslot, G. M.~J., Winands, M.~H., {Van Den Herik}, H.~J., Uiterwijk, J.~W.,
  Bouzy, B., et~al. (2008).
\newblock {Progressive Strategies For Monte-Carlo Tree Search}.
\newblock {\em New Mathematics and Natural Computation (NMNC)}, 4(03):343--357.

\bibitem[Cou{\"e}toux et~al., 2011]{couetoux2011continuous}
Cou{\"e}toux, A., Hoock, J.-B., Sokolovska, N., Teytaud, O., and Bonnard, N.
  (2011).
\newblock {Continuous upper confidence trees}.
\newblock In {\em {International Conference on Learning and Intelligent
  Optimization}}, pages 433--445. Springer.

\bibitem[Coulom, 2006]{coulom2006efficient}
Coulom, R. (2006).
\newblock {Efficient selectivity and backup operators in Monte-Carlo tree
  search}.
\newblock In {\em {International conference on computers and games}}, pages
  72--83. Springer.

\bibitem[Coulom, 2007]{coulom2007computing}
Coulom, R. (2007).
\newblock {Computing elo ratings of move patterns in the game of go}.
\newblock In {\em {Computer games workshop}}.

\bibitem[Haarnoja et~al., 2018]{haarnoja2018soft}
Haarnoja, T., Zhou, A., Abbeel, P., and Levine, S. (2018).
\newblock {Soft Actor-Critic: Off-Policy Maximum Entropy Deep Reinforcement
  Learning with a Stochastic Actor}.
\newblock {\em arXiv preprint arXiv:1801.01290}.

\bibitem[Kocsis and Szepesv{\'a}ri, 2006]{kocsis2006bandit}
Kocsis, L. and Szepesv{\'a}ri, C. (2006).
\newblock {Bandit based monte-carlo planning}.
\newblock In {\em {ECML}}, volume~6, pages 282--293. Springer.

\bibitem[Michalowicz et~al., 2013]{michalowicz2013handbook}
Michalowicz, J.~V., Nichols, J.~M., and Bucholtz, F. (2013).
\newblock {\em {Handbook of differential entropy}}.
\newblock Crc Press.

\bibitem[Moerland et~al., 2017]{moerland2017efficient}
Moerland, T.~M., Broekens, J., and Jonker, C.~M. (2017).
\newblock {Efficient exploration with Double Uncertain Value Networks}.
\newblock {\em Deep Reinforcement Learning Symposium @ NIPS 2017}.
\newblock arXiv preprint arXiv:1711.10789.

\bibitem[Moerland et~al., 2018]{moerland2018monte}
Moerland, T.~M., Broekens, J., Plaat, A., and Jonker, C.~M. (2018).
\newblock {Monte Carlo Tree Search for Asymmetric Trees}.
\newblock {\em arXiv preprint arXiv:1805.09218}.

\bibitem[Rosin, 2011]{rosin2011multi}
Rosin, C.~D. (2011).
\newblock {Multi-armed bandits with episode context}.
\newblock {\em Annals of Mathematics and Artificial Intelligence},
  61(3):203--230.

\bibitem[Silver et~al., 2016]{silver2016mastering}
Silver, D., Huang, A., Maddison, C.~J., Guez, A., Sifre, L., {Van Den
  Driessche}, G., Schrittwieser, J., Antonoglou, I., Panneershelvam, V.,
  Lanctot, M., et~al. (2016).
\newblock {Mastering the game of Go with deep neural networks and tree search}.
\newblock {\em Nature}, 529(7587):484--489.

\bibitem[Silver et~al., 2017a]{silver2017mastering2}
Silver, D., Hubert, T., Schrittwieser, J., Antonoglou, I., Lai, M., Guez, A.,
  Lanctot, M., Sifre, L., Kumaran, D., Graepel, T., et~al. (2017a).
\newblock {Mastering Chess and Shogi by Self-Play with a General Reinforcement
  Learning Algorithm}.
\newblock {\em arXiv preprint arXiv:1712.01815}.

\bibitem[Silver et~al., 2017b]{silver2017mastering}
Silver, D., Schrittwieser, J., Simonyan, K., Antonoglou, I., Huang, A., Guez,
  A., Hubert, T., Baker, L., Lai, M., Bolton, A., et~al. (2017b).
\newblock {Mastering the game of go without human knowledge}.
\newblock {\em Nature}, 550(7676):354.

\bibitem[Sutton and Barto, 2018]{sutton2018reinforcement}
Sutton, R.~S. and Barto, A.~G. (2018).
\newblock {\em {Reinforcement learning: An Introduction}}.
\newblock MIT press Cambridge, second edition.

\bibitem[Szegedy et~al., 2015]{szegedy2015going}
Szegedy, C., Liu, W., Jia, Y., Sermanet, P., Reed, S., Anguelov, D., Erhan, D.,
  Vanhoucke, V., and Rabinovich, A. (2015).
\newblock {Going deeper with convolutions}.
\newblock In {\em {2015 IEEE Conference on Computer Vision and Pattern
  Recognition (CVPR)}}.

\bibitem[Todorov et~al., 2012]{todorov2012mujoco}
Todorov, E., Erez, T., and Tassa, Y. (2012).
\newblock {Mujoco: A physics engine for model-based control}.
\newblock In {\em {Intelligent Robots and Systems (IROS), 2012 IEEE/RSJ
  International Conference on}}, pages 5026--5033. IEEE.

\bibitem[Williams, 1992]{williams1992simple}
Williams, R.~J. (1992).
\newblock {Simple statistical gradient-following algorithms for connectionist
  reinforcement learning}.
\newblock In {\em {Reinforcement Learning}}, pages 5--32. Springer.

\end{thebibliography}

\clearpage
\appendix
\section{Enforcing Action Space Bounds with Transformed Beta Distributions} \label{sec_bounds}
Continuous action spaces are generally bounded, i.e., we want to sample $\bm{a} \in [-c_b,c_b]^{n_a}$ for some constant $c_b \in \mathbb{R}^+$ and action space dimensionality $n_a$. There are various probability distributions with support on a continuous bounded interval. A well-known and flexible option is the Beta distribution, which has support in $[0,1]$. We will therefore make our network predict the parameters of a factorized Beta distribution $\bm{u} \sim q(\bm{u}) $, where each element $u_i \sim \text{Beta}(\alpha_i(\phi),\beta_i(\phi))$. Our goal is to transform the random variable $\bm{u}$ to a random variable $\bm{a}$ with support $\bm{a} \in [-c_b,c_b]^{n_a}$. A simple transformation $g$ that achieves this goal is
 
\begin{equation}
\bm{a} = g(\bm{u}) = c_b \cdot (2\bm{u} - 1) \label{eq_gu}
\end{equation}

For the loss specification in the paper, we require the (log)-density $\pi(\bm{a})$ of the transformed variable. We know from the change of variables rule that:

\begin{equation}
\pi(\bm{a}) = q(\bm{u}) \Big | \det(\frac{\mathrm{d} \bm{a}}{\mathrm{d} \bm{u}})  \Big |^{-1} 
\end{equation}

 For the transformation $\bm{a} = g(\bm{u})$, the Jacobian $\frac{\mathrm{d} \bm{a}}{\mathrm{d} \bm{u}} =  \mathrm{diag}(2c_b)$ is a diagonal matrix. Therefore, we can derive a simple expression for the (log-)likelihood of $\bm{a}$:

\begin{equation}
\pi(\bm{a}) = q(\bm{u}) \cdot (2c_b)^{-n_a}  , \quad \text{and} \quad \log \pi(\bm{a}) = \log q(\bm{u}). - n_a \cdot \log(2c_b).
\end{equation}

\subsection{Entropy of Transformed Beta Distribution} \label{sec_entropy_beta}
We know the entropy of a linear transformation of some variable from differential entropy \citep{michalowicz2013handbook}. For a linear transformation $\bm{M} \bm{u} + \bm{l}$, with matrix $\bm{M}$ and vector $\bm{l}$, we have

\begin{equation}
H(\bm{M} \bm{u} + \bm{l}) = H(\bm{u}) + \log | \det(\bm{M}) |
\end{equation} 

For our transformation $g(\bm{u})$ (Eq. \ref{eq_gu}), the second term of this equation equals $n_a \log(2c_b)$. Since this term does not depend on $\phi$, and therefore does not contribute any gradients, we will simply ignore it. The entropy of the Beta distribution $q(\bm{u})$ can be computed analytically (\citet{michalowicz2013handbook}, p.63). 

\end{document}